\documentclass{article} % For LaTeX2e
\usepackage{iclr2024_workshop_realign,times}

\usepackage{amsmath,amsfonts,bm}

\def\eqref#1{equation~\ref{#1}}
\def\1{\bm{1}}

\def\mC{{\bm{C}}}

\def\mH{{\bm{H}}}

\def\mZ{{\bm{Z}}}

\DeclareMathAlphabet{\mathsfit}{\encodingdefault}{\sfdefault}{m}{sl}
\SetMathAlphabet{\mathsfit}{bold}{\encodingdefault}{\sfdefault}{bx}{n}

\def\emH{{H}}

\usepackage{hyperref}
\usepackage{url}

\usepackage{graphicx}
\usepackage{multirow}
\usepackage{fancyhdr}
\usepackage{booktabs}
\usepackage{subcaption}
\usepackage{algorithm}
\usepackage{algpseudocode}
\usepackage{gensymb}
\usepackage{amsmath}
\usepackage{comment}
\usepackage{booktabs}
\usepackage{multirow}
\usepackage{xcolor}

\title{Explaining Human Comparisons using Alignment-Importance Heatmaps}

\author{Nhut Truong, Dario Pesenti \& Uri Hasson\thanks{Corresponding Author. ORCID  https://orcid.org/0000-0002-8530-5051} \\
Center for Mind/Brain Sciences (CIMeC)\\
University of Trento \\
Rovereto, Trento 38068, Italy \\
\texttt{\{leminhnhut.truong, dario.pesenti, uri.hasson \}@unitn.it} \\
}

\iclrfinalcopy % Uncomment for camera-ready version, but NOT for submission.

\begin{document}

\maketitle

\begin{abstract}
We present a computational explainability approach for human comparison tasks, using Alignment Importance Score (AIS) heatmaps derived from deep-vision models. The AIS reflects a feature-map's unique contribution to the alignment between Deep Neural Network's (DNN) representational geometry and that of humans.  We first validate the AIS by showing that prediction of out-of-sample human similarity judgments is improved when constructing representations using only higher-scoring AIS feature maps identified from a training set. We then compute image-specific heatmaps that visually indicate the areas that correspond to feature-maps with higher AIS scores. These maps provide an intuitive explanation of which image areas are more important when it is compared to other images in a cohort. We observe a correspondence between these heatmaps and saliency maps produced by a gaze-prediction model. However, in some cases, meaningful differences emerge, as the dimensions relevant for comparison are not necessarily the most visually salient. To conclude, Alignment Importance improves prediction of human similarity judgments from DNN embeddings, and provides interpretable insights into the relevant information in image space.
\end{abstract}

\section{Introduction}
\subsection{The question: Explaining human comparisons}
Work in recent years has shown that DNNs learn feature spaces whose geometry has some similarity to that of humans. This is convincingly shown by the fact that human similarity judgments (HSJs) for pairs of words or images are often quite well predicted by the distances between image-pairs or word-pairs in vision-DNNs or language models \citep[for reviews, see][]{battleday2021convolutional, roads2024modeling, sucholutsky2023getting}. These models therefore naturally extract features relevant for modeling HSJs when trained on standard tasks such as image classification or word prediction. While the object-embeddings of such pretrained machine learning models approximate HSJs quite well, it has been further shown that these predictions can be considerably improved using down-stream operations. 

One such operation is to learn a reweighting of the products of feature values, which  improves prediction of HSJs for both images \cite[e.g.,][]{peterson2018evaluating, kaniuth2022feature} and words \cite[e.g.,][]{Richie2021}. Another approach is to use supervised pruning to assess features' importance in the context of estimating a set of similarity judgments \citep{tarigopula2023improved, flechas2023enhancing}. Pruning does not alter the activation weights of the retained features, but instead removes a subset of features from the embedding matrix. Pruning has also been used to identify sub-spaces in language models that optimize particular classification tasks \cite[e.g.,][]{cao2021low}.

While prior work has shown that pruning of nodes in a DNN's penultimate layer can improve prediction of similarity judgments, here we are interested in its potential to explain what parts of an image matter for the judgment itself. Understanding which information is used as a basis for comparison is a fundamental question in cognitive science. Since the work of \citet{tversky1977features}, many studies have shown that comparisons between objects are a function of those elements that are shared or distinct between them. However, for naturalistic stimuli, it is difficult to know which properties are important when an image is compared to a target set of images. Here we suggest that this question is tractable via a computational solution in which latent dimensions that are related to the comparison process are identified and projected onto the image space as a heatmap. We release our code at \textit{https://github.com/tlmnhut/ais\_heatmap}

\subsection{Logic of the current study} 

We present the logic here, with a complete formal presentation provided in Section \ref{sec:prelim}. Our approach relies on evaluating how pruning changes the alignment between human and computer-model representational spaces. Both spaces are operationalized using pairwise distances between images. One set of distances is derived from human behavior ($HB_{dist}$), the other is computed from a computer model ($Model_{dist}$).  We define the baseline isomorphism between the two spaces as the correlation between these two vectors. 

In the next step, a perturbation is introduced to the feature representations of an image. Specifically, a feature map in the last convolutional layer is masked. Therefore, the information from that feature map is not encoded in the model, and not propagated onwards to the fully connected layer from which we obtain image embeddings. Subsequently, $Model_{dist}$ is recomputed, as is the isomorphism between the representations. Note that only the target image is affected, and not the other images. Furthermore, $HB_{dist}$ remains unchanged. There are two possible outcomes: \textit{i}) if the encoded information from the feature map is cognitively irrelevant or even confounding, its removal could alter $Model_{dist}$ in a way that improves the isomorphism with human similarity judgments. Conversely, \textit{ii}) if the encoded information from the feature map is cognitively-relevant (e.g., masking a feature map representing an animal’s face in context of similarity judgments between animals), its removal will alter $Model_{dist}$ in a way that decreases the isomorphism with human judgments. This occurs because the way that images stand in relation to each other in the DNN representation is now lacking information that underlies human judgments. By iteratively masking all feature maps in the last convolutional layer, each feature map is linked with a perturbation score indicating its importance.

Similar logic was presented in the previous works, but masking was applied on the image space rather than the latent feature space. For instance, \citet{tarigopula2023improved} used this approach with human neuroimaging data to explain which parts of an image contain information relevant to the representational space of various brain regions. In other work, \citet{palazzo2020decoding} masked image patches to evaluate how masking impacted the compatibility between vision-DNN embeddings and EEG data.

\subsection{Current aims and contribution} 

The current study's aims advances over prior studies in three respects: it directly studies human comparison processes, it introduces an advantageous masking procedure, and it evaluates the results against typical saliency maps. The aforementioned studies operationalized representational spaces from multivariate fMRI and EEG recordings but have not studied human comparison processes. Furthermore, the technique they use, namely, mask-sweep over an image, presents several major limitations: 1) the mask size is arbitrary, requiring the use of multiple sizes; 2) an arbitrary decision is required regarding how to combine information from different mask sizes; 3) the process is computationally costly, as masks are ideally applied with each pixel being in the mask center; 4) a theoretical weakness is that the mask is not informed by prior information contained in the model. 

Departing from these prior studies, here we directly model human comparison judgments, and use a different, more efficient approach to masking images, which uses information already present in the DNNs own feature space. Specifically, we focus on the feature maps in a deep convolutional layer, and use them to define the masks.  Our approach is inspired by Score-CAM \cite[Score-weighted Class Activation Maps;][]{wang2020scorecam} which is an explanatory method that generates heatmaps indicating which sections of a target image are relevant for its classification. Score-CAM takes the information in each feature map, upscales it to the original input resolution, uses it as an information selector for the original input image, and computes the activation for correct class (pre-softmax confidence) when using that feature map alone. After repeating this process for all feature maps, the confidence scores are used as weights to generate a heatmap highlighting image areas important for classification.  Using a similar logic, we show that information at the feature-map level is also highly useful for identifying which feature maps are important for the alignment between the DNN and human representational spaces, and that these can be visualized in a similar manner. 

Beyond our main explainability objective, we have two other important aims. First, we evaluate whether it is possible to identify feature maps that are particularly important for predicting human representational spaces; using only these feature maps should improve out of sample prediction accuracy for human similarity judgments as compared to using all feature maps.  Second, we evaluate the relationship between heatmaps produced using this method, and traditional saliency maps. While the latter operationalizes saliency using information latent in the image itself, the heatmaps we produce highlight information pertinent to image comparisons within a given set.

\section{Methods}
\subsection{Preliminaries} 
\label{sec:prelim}
\textbullet \hspace{3 pt} Architecture and datasets: In the main analysis, We use VGG-16, a deep neural network \citep{simonyan2014very}, pre-trained on ImageNet \citep{deng2009imagenet} and another trained on Ecoset\footnote{Available at https://osf.io/kzxfg/} \citep{mehrer2021ecoset}. VGG-16 was used because Ecoset was trained on that model. It is also a common architecture used for predicting human similarity judgements \citep{peterson2018evaluating, kaniuth2022feature} and has been shown to be a good candidate for behavior or brain alignment \citep{schrimpf2018brain}. As images we used a dataset provided by \citet{peterson2018evaluating}, which consists of 720 images divided into six categories of 120 images. The categories were: Animals, Fruits, Furniture, Various, Vegetables and Automobiles (the latter effectively including any means of transportation including horses, sleds, cranes; Transportation henceforth). Images had a native resolution of $500\times500$ which was downscaled to $224\times224$ to fit the model.

\noindent \textbullet \hspace{3 pt} Human Similarity Judgments: Let $\displaystyle \mH$ be a matrix representing the similarity judgments provided by human assessors for $n$ objects. Each entry $\displaystyle \emH_{i,j}$ in the matrix corresponds to the similarity judgment between objects $i$ and $j$. We use the upper triangle of matrix $\displaystyle \mH$, denoted as $\displaystyle \mH_u$. % The upper triangle is obtained by considering only elements $H_{ij}$ where $i < j$, excluding the diagonal elements $S_{ii}$.

\noindent \textbullet \hspace{3 pt} Object distances in feature space: Let $\displaystyle \mC$ be a matrix representing the embeddings of $n$ images onto $d$ features of the penultimate layer of the pre-trained computer vision model, denoted as $\displaystyle \mC \in \mathbb{R}^{n \times d}$. Specifically, we use VGG-16 with $d=4096$, and the number of images in each Peterson's category is $n = 120$. Matrix $\displaystyle \mC$ is obtained by considering all parameters of the pre-trained model, and specifically all 512 feature maps of the deepest convolutional layer. $\displaystyle \mZ_u$ is the upper triangle
of image-pair similarity matrix $\displaystyle \mZ$, computed from the Spearman correlation for each row pair in $\displaystyle \mC$.

\noindent \textbullet \hspace{3 pt} Subspaces in matrix $\displaystyle \mC$: We produce two variants of $\displaystyle \mC$ (all with dimension $n \times d$). The first variant (``remove 1''), denoted as $\displaystyle \mC^{(\neg k)}$, is constructed by excluding feature map $k$ where  \(k \in \{1, 2, \ldots, 512\}\). The second variant is produced when using only a subset $S$ of feature maps in the model. Let \(S \subseteq \{1, 2, \ldots, 512\}\) be a set of selected feature-map indices, and let $\displaystyle \mC^{(S)}$ be the matrix representing the embedding of \(n\) images onto $d$ nodes in the penultimate layer, but when using the subset of feature-maps corresponding to \(S\). Note that in all cases, the (one or more) feature-map activations are propagated to the penultimate layer using the pre-trained weights.

\noindent \textbullet \hspace{3 pt} From the variants of $\displaystyle \mC$ we derive matching similarity matrices. The first, $\displaystyle \mZ^{(\neg k)}$, is obtained by computing the cosine similarity for each pair of rows in $\displaystyle \mC^{(\neg k)}$. The second, $\displaystyle \mZ^{(S)}$ is formed using the selected feature indices in $\displaystyle \mC^{(S)}$.

\noindent \textbullet \hspace{3 pt} As indicated, $\displaystyle \mZ_u$ and $\displaystyle \mH_u$ denote the vectorized upper triangles of matrices $\displaystyle \mZ$ and $\displaystyle \mH$ respectively. The Spearman  correlation coefficient between the two is denoted as $\rho({\displaystyle \mZ_u, \displaystyle \mH_u})$. We refer to this value as a Baseline Second-Order-Isomorphism (2OI) between the two domains. Analogously, in some cases we compute $\rho({\displaystyle \mZ^{(\neg k)}_{u}, \displaystyle \mH_u})$ and $\rho({\displaystyle \mZ^{(S)}_{u}, \displaystyle \mH_u})$.
    
\subsection{Aim 1: Identifying a subset of feature maps that optimizes prediction of human similarity judgments}

We define the Alignment Importance Score (AIS) of each feature map in terms of its predictive capacity for the human representation $\displaystyle \mH_u$. Intuitively, we aim to determine how the removal of each feature map \(k \in \{1, 2, \ldots, 512\}\) affects the baseline isomorphism, $\rho({\displaystyle \mZ_u, \displaystyle \mH_u})$. The removal of each feature map produces a modified 2OI score, $\rho({\displaystyle \mZ^{(\neg k)}_{u}, \displaystyle \mH_u})$.  Finally, The AIS of feature map $k$ is defined in Equation \ref{eq:contri}, with positive values indicating a relatively important feature map, and negative values a less important one. After computing AIS for all feature-maps, we rank-order them based on their AIS.

\begin{equation}
    \text{AIS}_{k} = \rho({\displaystyle \mZ_u, \displaystyle \mH_u}) - \rho({\displaystyle \mZ^{(\neg k)}_{u}, \displaystyle \mH_u})
    \label{eq:contri}
\end{equation}

We then identify an optimal subset of feature maps for predicting \(\displaystyle \mH_{u}\). In each iteration, one feature map is added to the subset $S$ in descending order of AIS rank, and we recompute the 2OI, $\rho({\displaystyle \mZ^{(S)}_{u}, \displaystyle \mH_u})$ using that subset of feature maps alone. After these 512 iterations, subset $S^{*}$ ultimately selected is the one that maximizes 2OI.

To validate AIS, we use an 80:20 cross-validation framework where 80\% of the entries in \(\displaystyle \mH_u\) are assigned to a training set, and the remaining 20\% constitute the test set. The optimal subset of feature map indices, \(S^{*}\), is determined from the training set using sequential features selection as described above. For statistical significance testing, we repeat the entire cross-validation process eight times with different dataset shuffling. This produces 40 Full vs. Retained value-pairs for each relevant comparison. To evaluate generalization, we use only this \(S^{*}\) set of feature maps to predict HSJs on the test set. Prediction performance is compared against a baseline where all 512 features are used for predicting HSJs in the test set. Statistical significance testing, per dataset, is based on the 40 value-pairs produced via cross-validation, which are analyzed using paired two-tailed T-tests (12 tests in all, non-corrected for multiple comparisons). Success of Aim 1 is determined if \(\rho(\displaystyle \mZ^{(S^*)}_{u}, \displaystyle \mH_u)\) surpasses \(\rho(\displaystyle \mZ_{u}, \displaystyle \mH_u)\), indicating superior prediction compared to the baseline using a subset of feature maps.

As an additional baseline, we used LPIPS \cite[Learned Perceptual Image Patch Similarity,][]{zhang2018unreasonable}, which is a method for obtaining a cognitively-relevant similarity metric between image pairs. LPIPS fine-tunes a  computer vision CNN so that the image distances in the network, calculated as differences between embedding vectors, align with human similarity judgments. LPIPS is fine-tuned using human decision data regarding which of two slightly altered images are closer to an original image, and is based on reweighting all layers of the network. LPIPS has shown to closely match human behavior in 2-Alternative Forced Choice tasks involving minor image distortions and a reference image. To evaluate whether LPIPS is at all viable for our materials and similarity judgments, we applied LPIPS to all images in each dataset to compute pairwise distances between images, and computed the Pearson correlation between the LPIPS distance matrix and the human similarity judgments. Note that the LPIPS method does not allow integration with pruning, as its reweighting function achieves a parallel goal. We use the pre-trained LPIPS weights provided by the original authors as these have been trained on a large set of human judgments and have been argued to predict human behavior in multiple domains.

\subsection{Aim 2: Explaining human similarity judgments}
\label{sec:explainSim}

Our goal is to identify which image patches, in image space, are relevant to comparisons between a target image $t$ and other images in the set. This is visualized by creating a heatmap for $t$ identifying those image sections, as follows. We begin by defining a baseline 2OI for $t$ as the Spearman correlation between the $n-1$ similarity judgments associated with $t$, as quantified from the model, and the corresponding set of human similarity judgments. As in Aim 1, we define the AIS of feature map \(k\) by computing a value that reflects the departure from baseline, as indicated in Equation \ref{eq:contri}. 

We iterate over all 512 feature maps, producing 512 AIS values that indicate the relative importance of each feature map for the alignment between DNN-derived distances and human similarity judgments for target image $t$. This produces an \(n \times k\) matrix (120 [AIS] x 512 [feature map]) for each dataset containing 120 images. We then compare these distributions between the ImageNet and Ecoset-trained models to understand if and how the training regime impacts the distribution of AIS. Histograms are computed for the mean AIS value by feature, and the Mean Absolute Deviation, computed by feature (column) and by image (row). 

Image-level heatmaps are then computed as follows. We first convert negative AIS values to zero because they indicate features that encode information less relevant to modeling the human data (see Eq. \ref{eq:contri}). The remaining scores are sum normalized. Subsequently, feature maps for an image are weighted-averaged according to their corresponding AIS to create a  heatmap. In the heatmaps, warmer colors indicate image areas associated with the more important features.

To quantify the similarity between the heatmaps generated by Ecoset and Imagenet, we defined a Match score for each image as the Pearson correlation between the heatmap generated by the Ecoset model and the one generated by the Imagenet model. Anticipating the results, in certain instances, the Match score was low. We therefore examined if this occurred for images that did not correspond to classes on which the models were trained. For each image, we computed the entropy of the post-softmax probability distributions, independently for the Ecoset and ImageNet trained models. The higher of these two entropy values was retained and designated as maxEntropy. Subsequently, considering all images in a dataset, we computed the correlation between the Match score and maxEntropy.

\subsection{Aim 3: Cross-referencing heatmaps against saliency maps}
We compare the heatmaps produced by our method to those produced by TranSalNet \cite{lou2022transalnet}, which is a state-of-the-art DNN that identifies salient image sections and accurately predicts human gaze patterns (see Figure \ref{fig:appendix_trans} in Appendix). We cross-reference TranSalNet against our method (AIS) using two approaches: Precision-Recall curves and Subset analyses.
\subsubsection{Precision-Recall curves}
First, we evaluate how well a pixel's salience predicts its inclusion in an AIS heatmap. When the salience and AIS maps are thresholded at a specified level to form binarized maps, the relationship between them can be understood in terms of precision and recall. The binarized AIS map is treated as the target variable, and the binarized saliency map is the predicting variable. In this case, we have:
\[
\text{Precision} = \frac{|TranSalNet \cap AIS|}{|TranSalNet|}
\] 
and 
\[
\text{Recall} = \frac{|TranSalNet \cap AIS|}{|AIS|}.
\] 
We describe this relationship using a Precision-Recall curve. The curve is generated by thresholding the AIS map at a fixed level and then plotting precision versus recall as the saliency map is thresholded across a range of levels.

The following steps were performed for each image: first, we created a heatmap as described in Aim 2 and generated a corresponding saliency map using TranSalNet. We kept the same aspect ratio of the images input to both VGG-16 and TranSalNet for compatibility in later comparisons. We conducted four separate analyses, where we created a binary mask for the AIS map at each of the following percentiles: \( P = \{60, 70, 80, 90\} \). In each analysis we thresholded the saliency maps at all percentiles between 1 and 99, with a step size of 2. Percentiles were calculated separately for each image.  

% changed the name of section from Subset analysis to conditional probability analysis.
\subsubsection{Conditional probability analysis}
In this analysis we aim to identify whether an image section (specifically, a pixel) identified as salient (\textit{Sal}) is more likely to also be identified as comparison-relevant (\textit{CR}; that is, warm-colored in our analysis). To do this we threshold both maps to select the top 5\% of Salient and \textit{CR} pixels, producing \textit{Sal}, $\neg Sal$, \textit{CR} and $\neg CR$ partitions of the image pixels. We then compute the Relative Risk (RR) ratio as in Equation \ref{eq:rr}. 

% \[ \text{RR} = P(CR | Sal) \div P(CR | \neg Sal) \]
\begin{equation}
    \text{RR} = P(CR | Sal) \div P(CR | \neg Sal)
    \label{eq:rr}
\end{equation}

The relative risk as computed here measures the likelihood  of \textit{Sal} pixels being \textit{CR} pixels compared to $\neg Sal$ pixels.  An $RR$ value greater than 1 indicates that salient pixels are more likely to be CR than non-salient ones, while an RR less than 1 indicates the opposite. A main difference between this analysis and the precision-recall one is that it also quantifies joint distributions within the non-salient pixel-set.  
We repeat this analyses when thresholding both maps at 10\% and 15\% top \textit{Sal} and \textit{CR} pixels. 

We note that there is no requirement that the two methods identify the same image features. The saliency map is driven by image features (including higher level semantics captured by the DNNs), whereas the heatmap we produce from AIS values is a function of how a certain object stands in relation to other objects in the set. As we will see, this produces cases of very high overlap, but also important distinctions.

\subsection{Aim 4: Generalization to other architectures and training objectives}

In Aims 1, 2 and 3 the image embeddings used were obtained from VGG-16. VGG-16, and a later variant VGG-19, are somewhat unique in that after the deepest convolutional layer, they also include two very large fully connected layers. These layers perform non-linear, abstract interactions over the information in the deepest feature map layer, and are essential for linking this information to the classification task.

Many other computer-vision architectures do not include such layers, and instead use the deepest feature maps, relatively directly, for classification. This is done by implementing global average pooling, which reduces each of these feature maps into a single value, followed by learning a linear combination of these values for classification. Thus, in these architectures, the final layer before classification receives an input corresponding to the number of feature maps (after global pooling), and produces an output corresponding to the number of classes to be learned. 

To evaluate the applicability of the AIS-based analysis to other architectures, we applied the analysis developed for Aim 1, with several modifications, to the following models: Inception-V3 \citep{InceptionV3}, ResNet-152 \citep{Resnet152}, DenseNet-161 \citep{Densenet161}, EfficientNet-B3 \citep{tan2019efficientnet}, RegNetY-400MF \citep{regnet400}, and ResNeXt-50-32x4d \citep{resnext50}. The deepest layers of these architectures contain varying numbers of feature maps: Inception-V3, ResNet-152, and ResNeXt-50-32x4d each have 2,048 feature maps, DenseNet-161 has 2,208, EfficientNet-B3 has 1,536, and RegNetY-400MF has 440. 

We note that all these architectures learn features in the context of supervised classification tasks. To evaluate feature maps produced by non-supervised learning, we used a ResNet-50 architecture trained with the Barlow Twins self-supervised learning framework \citep{zbontar2021barlow}. 
In this approach, the objective of the the model is learn representations by maximizing the similarity between two augmented versions of the same image. In this way, training extracts general visual features, ignoring small visual distortions.   

For each of these architectures we performed five-fold Cross validation, as detailed for Aim1. For all architectures except VGG-16 and VGG-19, object embeddings were generated by applying global pooling to the feature maps from the deepest convolutional layer. For VGG-16 and VGG-19, embeddings were constructed from the penultimate, fully connected layer.

\section{Results}
\subsection{Aim 1: Identifying a subset of feature maps that optimizes prediction of human similarity judgments}

% Nhut regenerated figure after shifting from pearson-pearsonR2 to cosine-spearman, and split 80:20 on the sim pairs, not on the images
\begin{figure}[tb] 
    \centering
    \includegraphics[width=0.9\linewidth]{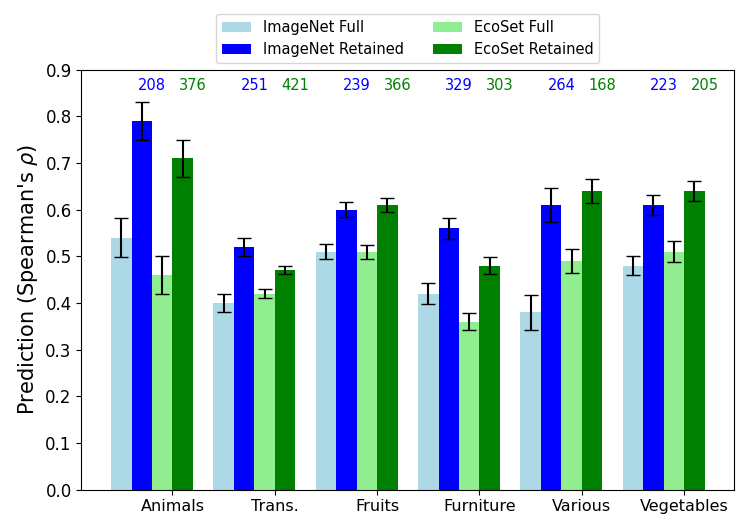} 
    \caption{Out-of-sample predictions of human similarity judgments using image embeddings. Full: using all 512 feature maps. Retained: using feature maps identified from an independent training set. The numbers above the second and fourth columns in each group represent averages of feature-map set sizes across 40 folds. Error bars indicate standard errors adjusted for paired-comparisons \citep{loftus1994using}.}
    \label{fig:aim_1_pruning}
\end{figure}

As shown in Figure \ref{fig:aim_1_pruning}, by computing AIS it was possible to identify a subset of 512 feature maps for each dataset, which produced improved out-of-sample predictions compared to a baseline condition where all feature maps were used. This was consistent for models trained on Ecoset or ImageNet, with less than 50\% of the 512 feature maps being used in 5/12 cases. Paired T-tests indicated that in all 12 cases, predictions from Full features were less accurate than those from features learned via pruning (p-values $<$ 0.01). The performance metrics of ImageNet and Ecoset were quite similar.

% Numbers update here August 2024 after shift to Spearman
Speaking to category-specific information, AIS values for each feature-map differed across datasets. That is, feature maps important for aligning one category were not necessarily important for another category. To evaluate this issue, we computed pair-wise Pearson correlations between the AIS values of the 512 feature-maps for each pair of datasets (e.g., Fruits vs. Vegetables). For both Ecoset and ImageNet, the strongest correlation was between Fruits and Vegetables (Ecoset $R = 0.48$; ImageNet $R = 0.67$). For Ecoset, the second highest correlation was between Transportation and Furniture ($R = 0.38$), whereas for ImageNet it was between Various and Animals ($R = 0.26$). Most of other correlations, in both analyses, ranged from -0.2 to 0.2.  

Finally, we evaluated the LPIPS method for human similarity modeling (see \textit{Methods}). LPIPS image-distances indeed tracked human similarity judgments for all categories, in that higher LPIPS distances were associated with lower similarity. However, these correlations were quite low. Spearman rho values were: Animals 0.15, Automobiles 0.19, Fruits 0.15, Furniture 0.07, Vegetables 0.40, and Various 0.19. Thus, alignment with LPIPS did not approach the levels seen in Figure \ref{fig:aim_1_pruning}, even for the non-pruned cases.

\subsection{Aim 2: Explaining human similarity judgments}
\begin{figure}[tb]
    \centering
    \includegraphics[width=\textwidth]{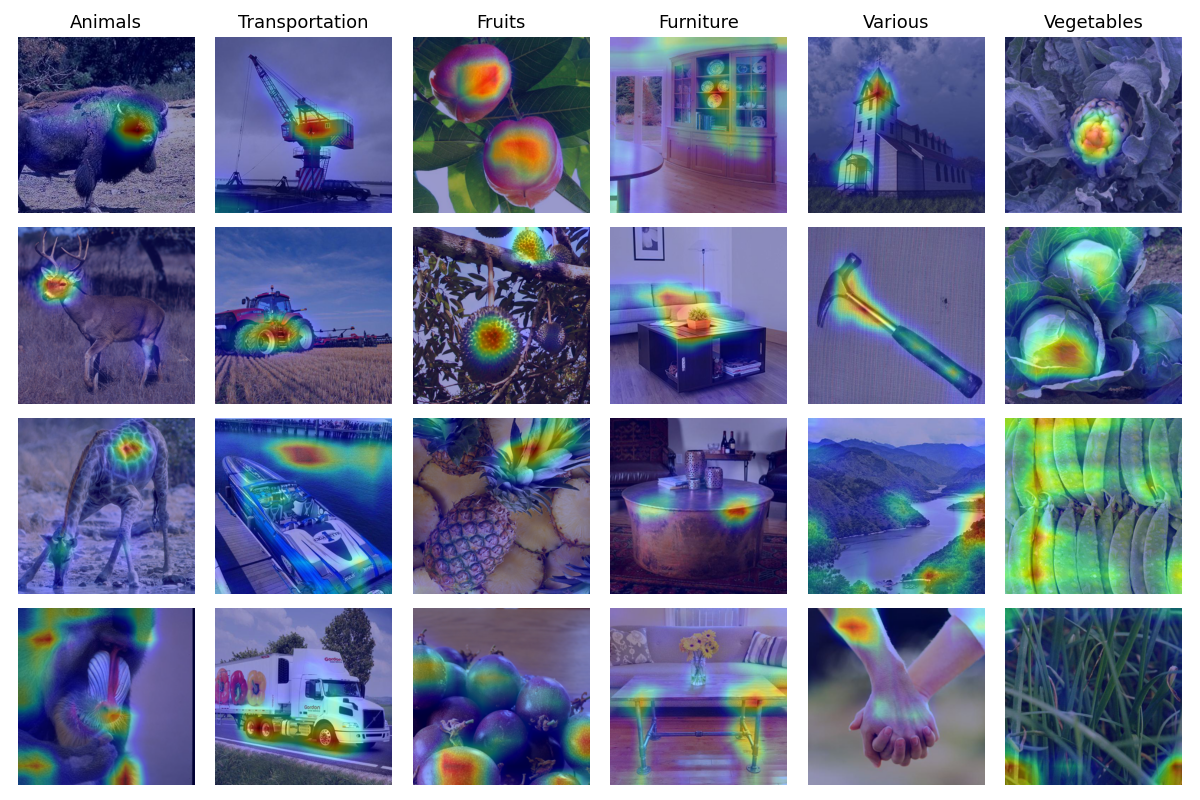}
    \caption{Heatmaps generated using Alignment Importance Scores of feature maps trained with Ecoset. For each dataset, two images with subjectively higher interpretability (top two rows) and lower interpretability (bottom two rows) were selected.}
    \label{fig:aim_2_heatmap}
\end{figure}

Figure \ref{fig:aim_2_heatmap} shows examples of heatmaps produced by alignment importance scoring. Given that each dataset contained 120 images, we selected 4 images from each dataset according to the principle that two of the images produced apparently sensible results, and the two others were less sensible. It can be seen that the method can identify image-sections that are relevant for inter-category comparisons, such as the faces of animals, central parts of fruits and vegetables, and discriminating elements of artifacts and man made objects. As we will see later, these are not necessarily the most salient aspects of images. 

To assess the similarity of heatmaps produced by Ecoset and ImageNet, for each image we calculated the correlation between the heatmaps produced by the two methods. The median  correlation values were as follows: $0.80\pm 0.16$ for Animals, $0.64\pm0.19$ for Transportation, $0.73\pm0.22$ for Fruits, $0.64\pm0.22$ for Furniture, $0.64\pm0.27$ for Various, and $0.56\pm0.25$ for Vegetables. In all datasets the maximum correlation values approached 1.0, while the minimum values often approached zero (see the histogram in Appendix Figure \ref{fig:aim_2_hist_heatmap}). As Appendix Figure \ref{fig:aim_2_hist_heatmap} shows, for all categories (apart from Animals), around 10\% of images showed a low correlation of less than 0.2. Considering a correlation of 0.8 as an (arbitrary) reference point for strong correspondence between heatmaps, we find that for Animals more than 40\% of the images showed correlations that exceeded this value, whereas for Transportation and Furniture the value was below 20\%. 

This means that although agreement was often good, training models on Ecoset or ImageNet often produces different heatmaps. These findings are consistent with those of Aim 1, which showed that the VGG-16 models trained on the two datasets capture and learn human similarity judgments in slightly different ways.  

As detailed section \ref{sec:explainSim}, we evaluated if images that presented a lower Match between Ecoset and ImageNet heatmaps were associated with higher entropy of post-softmax values in either of the two sets (maxEntropy), which would produce a negative correlation between the two quantities. We found that this was indeed the case, for Animals ($R = -0.31$), Fruits ($R = -0.34$), Various ($R = -0.24$), and Vegetables ($R = -0.21$). Weaker, yet sill negative correlations were found for Transportation and  -0.11, Furniture, $Rs = -0.11,  -0.04$ respectively. Thus, images that do not present information sufficient for classification produce disagreement between the two models. These might be out of distribution images or bad examples of trained categories. 

Ultimately, in those cases where heatmaps differ, the results of Aim 1 may be used as a guide to inform whether Ecoset or ImageNet is more plausible with respect to the human representation of a given category. For instance, given the low agreement in heatmaps produced for Transportation and Furniture, one may select to use the ImageNet produced feature maps as these provide better out-of-sample prediction of human behavior. 

\begin{figure}[tb]
  \centering

  \begin{subfigure}{\textwidth}
    \centering
    \includegraphics[width=\textwidth]{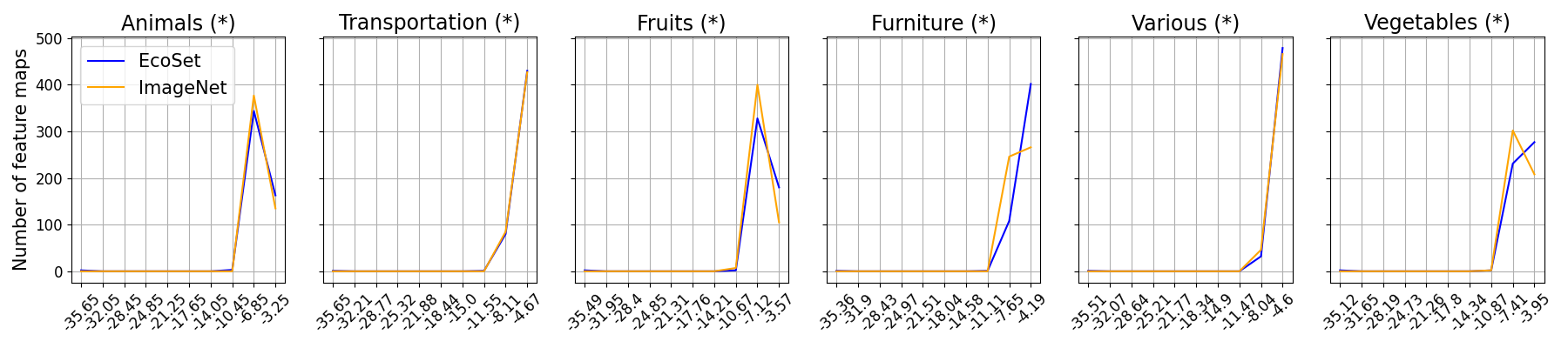}
    \caption{Log average of AIS values per feature.}
    \label{fig:avghist}
  \end{subfigure}

  \begin{subfigure}{\textwidth}
    \centering
    \includegraphics[width=\textwidth]{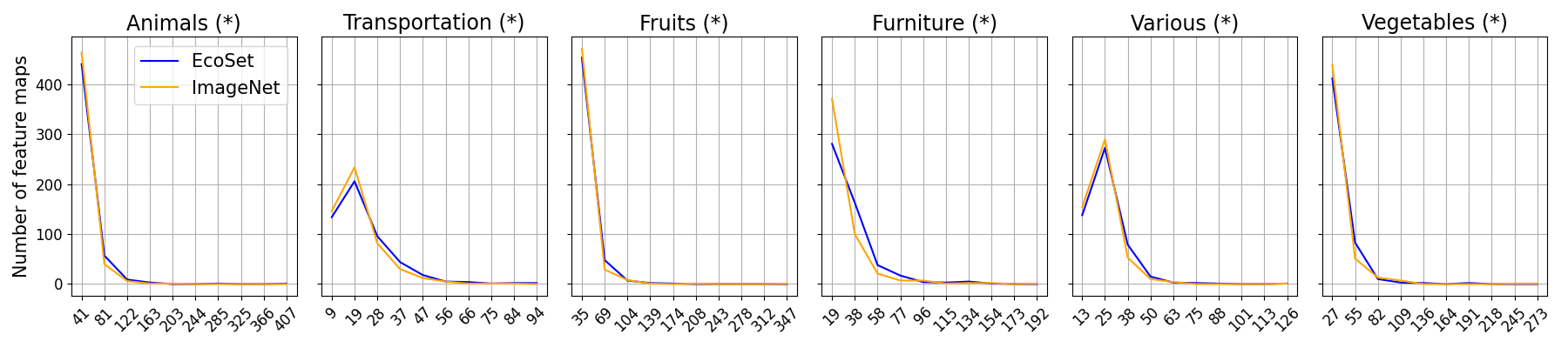}
    \caption{Mean absolute deviation of each feature's AIS values, computed over 120 images.}
    \label{fig:madcol}
  \end{subfigure}

  \begin{subfigure}{\textwidth}
    \centering
    \includegraphics[width=\textwidth]{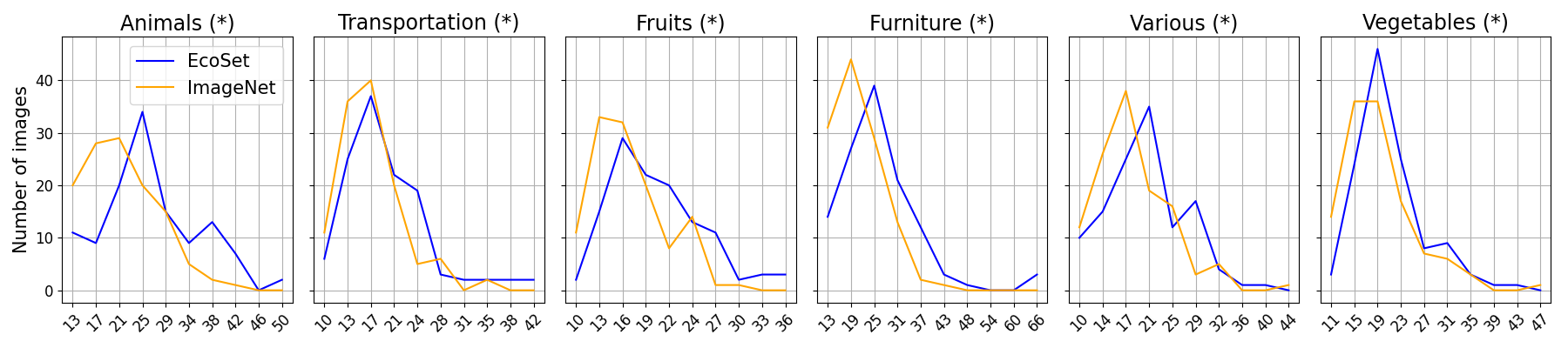}
    \caption{Mean absolute deviation of each image's AIS values, computed over 512 feature maps.}
    \label{fig:madrow}
  \end{subfigure}

  \caption{Histograms describing statistics of Alignment Importance Score distributions for models trained on Ecoset or ImageNet. The x-axis of (b) and (c) are displayed in e-4 format. A star symbol (*) indicates a significant difference between the two distributions as determined by a KS test.}
  \label{fig:aim_2_hist}
\end{figure}

% after conversion to spearman rho AIS mas, all comparisons in 3 sub-figures are significantly different, all p values are lower than 0.05
We also statistically quantified the relation between AIS values obtained for feature maps when produced from models trained on Ecoset or ImageNet. Figure \ref{fig:aim_2_hist} shows, for each dataset, histograms computing the Average AIS associated with each feature (log10 scaled), and the Mean Absolute Deviation computed per feature (column) and per image (row). The histogram shows that the average AIS rarely exceeded 0.001 for any feature (Figure \ref{fig:avghist}). Two-sided Kolmogorov-Smirnov (KS) tests \citep{hodges1958significance} were conducted to verify if the histograms associated with the two training regimes (ImageNet, Ecoset) came from the same distribution. Overall, KS test confirmed significant differences for all six categories ($ p<.05$).
  
With respect to Mean Absolute Deviation (MAD), when computed per feature (Figure \ref{fig:madcol}) we find that the values varied around one order of magnitude, with a few features showing relatively higher values meaning they were much more important for some images than others. The MAD histograms computed from per-image data indicated that ImageNet's AIS distribution was consistently left shifted with respect to Ecoset's (Figure \ref{fig:madrow}). This means that the AIS produced by Ecoset-trained model are  more strongly distributed, suggesting a more meaningful separation between those features relevant for alignment and those that are not. Two-sided Kolmogorov-Smirnov tests on Mean Absolute Deviation verify significant differences between the two models in all cases (all six datasets, KS tests, $ p<.05$). 

% histograms

% However, the correlations for Transportation (-0.06) and Furniture (0.02) do not follow this pattern.

\subsection{Aim 3: Cross-referencing heatmaps against saliency maps}
% commmented out all centered aligned table

\subsubsection{Precision-Recall curves}
For each image, we thresholded the AIS-produced heatmap at a given threshold to form a binary prediction target with AIS-related image sections (after thresholding) constituting the positive class. We then evaluated the extent to which these could be predicted by the saliency maps, using a Precision-Recall curve.  In this analysis, the target variable is thresholded at a fixed level (e.g., 90th percentile), while the predicting variable is thresholded across a range of levels, with precision and recall computed for each threshold. 

\begin{figure}[tb]
    \centering
    \begin{minipage}{0.50\textwidth}
        \centering
        \includegraphics[width=\textwidth]{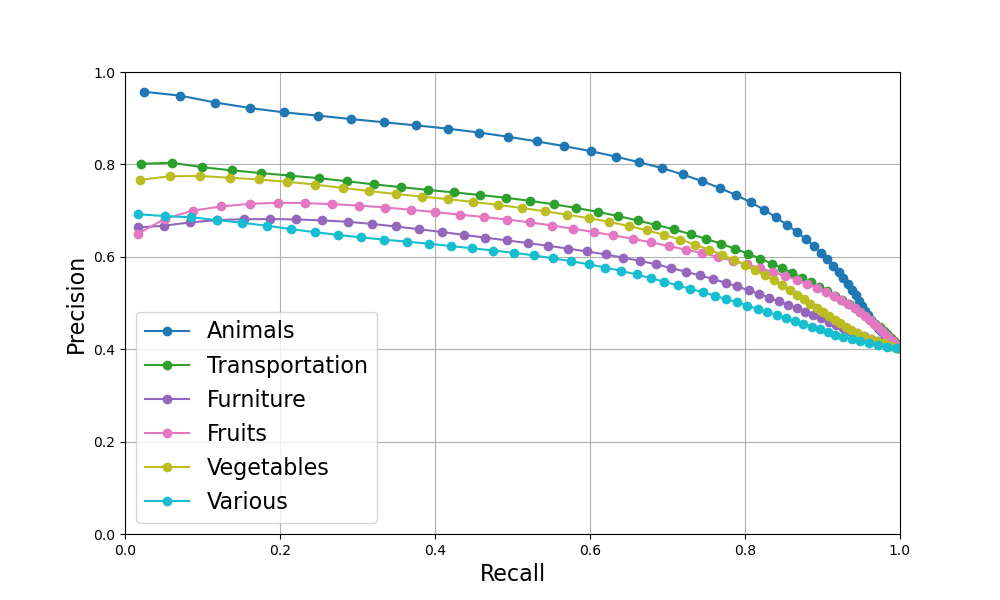}
        \subcaption{AIS heatmaps thresholded at 60th percentile}
        \label{fig:pr_curve_60}
    \end{minipage}\hfill
    \begin{minipage}{0.50\textwidth}
        \centering
        \includegraphics[width=\textwidth]{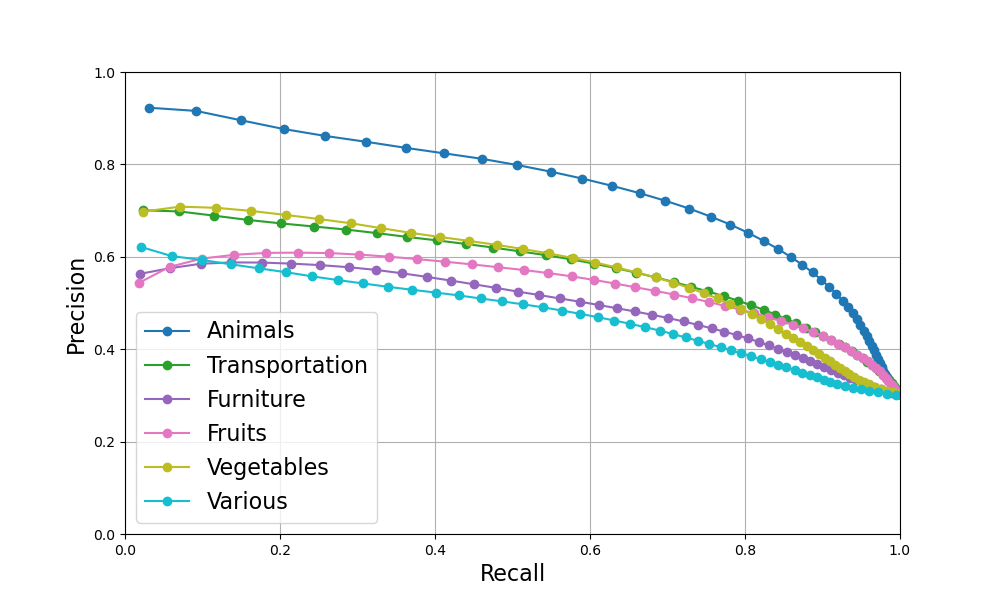}
        \subcaption{AIS heatmaps thresholded at 70th percentile}
        \label{fig:pr_curve_70}
    \end{minipage}
    
    \vspace{0.5cm}
    
    \begin{minipage}{0.50\textwidth}
        \centering
        \includegraphics[width=\textwidth]{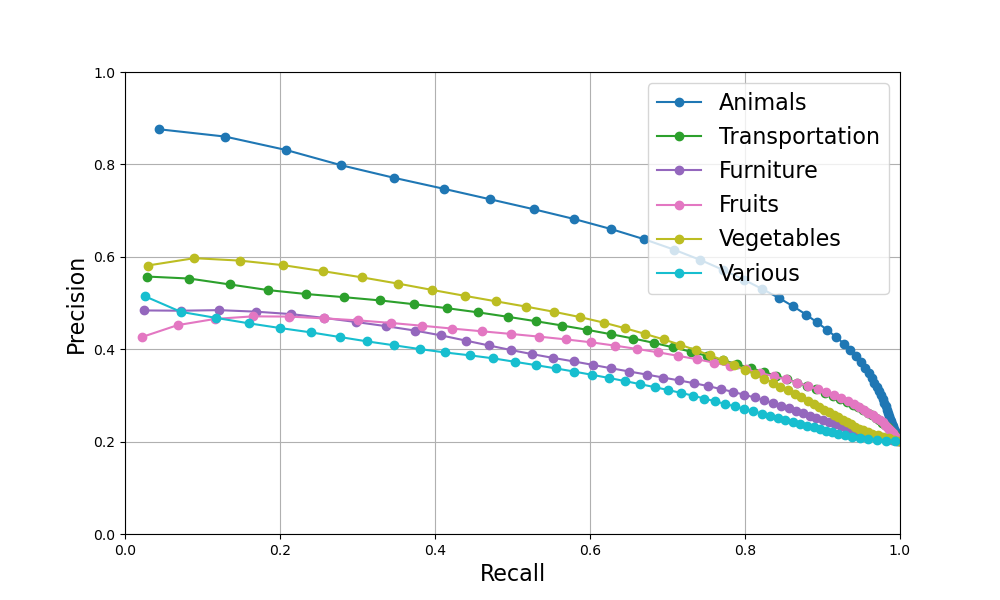}
        \subcaption{AIS heatmaps thresholded at 80th percentile}
        \label{fig:pr_curve_80}
    \end{minipage}\hfill
    \begin{minipage}{0.50\textwidth}
        \centering
        \includegraphics[width=\textwidth]{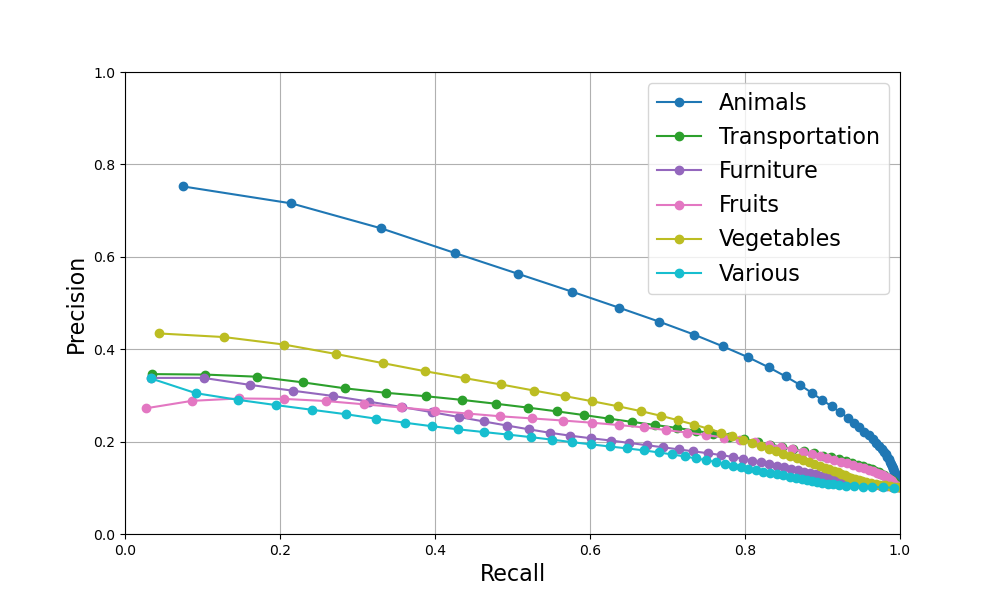}
        \subcaption{AIS heatmaps thresholded at 90th percentile}
        \label{fig:pr_curve_90}
    \end{minipage}

    \caption{Precision-Recall Curves when predicting AIS heatmap values from saliency, for different thresholds of AIS heatmaps. The target variable was heatmap values produced from AIS scores computed from ImageNet training. The predicting variable were saliency map values obtained from TranSalNet.}
    \label{fig:pr_curves}
\end{figure}

Figure \ref{fig:pr_curve_60} shows the results when predicting AIS heatmaps produced from ImageNet-trained feature maps, and with the AIS heatmaps thresholded at the 60th percentile. We observe that when saliency maps are thresholded at stringent levels (leftward points on the curve), precision is high for the Animals category, and somewhat lower for other categories, with values ranging from 0.6 to 0.8.

Lowering the threshold increased recall, but also gradually lowered precision as expected, While saliency and AIS maps were clearly related, with the exception of Animals, predicting AIS  from saliency appeared limited, even when AIS heatmaps were thresholded at a relatively low value of 60th percentile. The other panels in Figure \ref{fig:pr_curves} show the same analysis with AIS heatmaps thresholded at the 70th, 80th, and 90th percentiles. In the latter analysis, AIS-relevant pixels are defined as the top 10\%, and as shown in Figure \ref{fig:pr_curve_90}), saliency predicted membership in this class poorly, with the exception of the Animals category. Another observation is that for the Fruits category, thresholding the saliency map at the most strict level (left-most point) did not produce the highest precision, which was instead achieved  at lower thresholds. This suggests that the most salient points were not always the most precise predictors of AIS heatmaps..

In summary, we found that, with the exception of the Animals category, saliency heatmaps could not predict AIS heatmaps with good precision and recall, particularly when AIS heatmaps were thresholded at higher levels. A very similar pattern was found for AIS heatmaps produced from Ecoset feature maps (see Appendix Figure \ref{fig:ecoset_pr_curves}).

\subsubsection{Conditional probability analysis}
We observed that areas identified as comparison-relevant by AIS heatmaps were much more likely to be associated with salient image sections than with non-salient image sections, as indicated by  Relative Risk values strongly exceeding 1.0 (Table \ref{tab:aim_3_rr_avg}). This was found regardless of whether pixels in both heatmaps were thresholded at top 5\%, top 10\% or top 15\%. As the table shows, the RR values often exceeded 5, reaching as high as 30 for Animals. The data were quite similar for ImageNet and Ecoset overall. Furthermore, the Relative Risk values varied significantly across categories, being highest for Animals, and lowest for Vegetables. This suggests that for Animals, elements salient in images are also important for comparison, whereas this is less so for Vegetables.  This is numerically consistent with the Precision-Recall analysis where we found that thresholding saliency maps at high percentiles produced good prediction-precision of AIS data.

\begin{table}[tb]
\caption{Relative Risk values comparing heatmaps computed from Alignment Importance Scores to those generated by TranSalNet, a saliency model that predicts human gaze. Chance values are $RR=1$.}
\label{tab:aim_3_rr_avg}
\centering
\resizebox{\textwidth}{!}{
\begin{tabular}{l*{6}{c}}
\toprule
\multirow{2}{*}{Category} & \multicolumn{3}{c}{Ecoset} & \multicolumn{3}{c}{ImageNet} \\
\cmidrule(lr){2-4} \cmidrule(lr){5-7}
& 5\% vs. 5\% & 10\% vs. 10\% & 15\% vs. 15\% & 5\% vs. 5\% & 10\% vs. 10\% & 15\% vs. 15\% \\
\midrule
Animals & 30.8 $\pm$ 32.1 & 17.0 $\pm$ 18.3 & 12.7 $\pm$ 11.0 & 28.2 $\pm$ 34.2 & 14.9 $\pm$ 11.5 & 11.4 $\pm$ 8.2 \\
Transportation & 7.8 $\pm$ 11.5 & 5.8 $\pm$ 7.0 & 5.2 $\pm$ 6.5 & 6.4 $\pm$ 7.8 & 5.6 $\pm$ 5.8 & 5.3 $\pm$ 5.2 \\
Fruits & 9.9 $\pm$ 18.5 & 7.4 $\pm$ 10.9 & 6.2 $\pm$ 9.2 & 9.9 $\pm$ 21.4 & 6.6 $\pm$ 11.4 & 5.4 $\pm$ 8.2 \\
Furniture & 6.1 $\pm$ 10.3 & 5.1 $\pm$ 6.2 & 4.5 $\pm$ 4.8 & 6.5 $\pm$ 12.0 & 5.2 $\pm$ 6.5 & 4.6 $\pm$ 4.5 \\
Various & 17.3 $\pm$ 27.4 & 10.2 $\pm$ 11.0 & 8.7 $\pm$ 8.9 & 14.4 $\pm$ 31.4 & 8.2 $\pm$ 9.9 & 6.7 $\pm$ 7.2 \\
Vegetables & 6.4 $\pm$ 10.7 & 4.9 $\pm$ 6.8 & 4.1 $\pm$ 4.2 & 7.1 $\pm$ 14.7 & 5.0 $\pm$ 6.8 & 4.1 $\pm$ 4.2 \\
\midrule
All datasets & 13.0 $\pm$ 22.1 & 8.4 $\pm$ 11.7 & 6.9 $\pm$ 8.4 & 12.1 $\pm$ 23.8 & 7.6 $\pm$ 9.6 & 6.2 $\pm$ 6.9 \\
\bottomrule
\end{tabular}
}
\end{table}

\begin{figure}[tb]
    \centering
    \includegraphics[width=\linewidth]{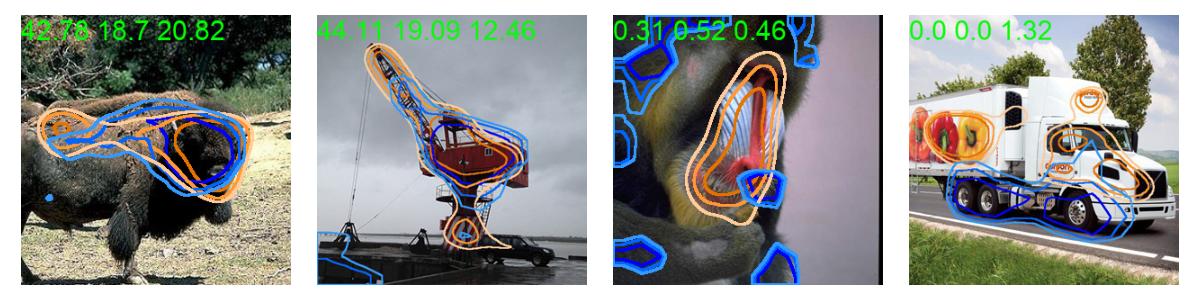}
    \caption{Overlap between the heatmaps created by Alignment Importance Scores (blue contours) and the saliency maps from TranSalNet (orange contours). The contours indicate the 5\%, 10\%, and 15\% most important pixels, with increasing color intensity respectively. Relative Risk values computed from  top 5\%, 10\% and 15\% pixels in each map are printed on the top of each images. The two left images are examples of cases where AIS and saliency identified similar areas, whereas the two right images present extreme cases of non-overlap. %For example, the TranSalNet model predicts human gaze on the cabin and the advertisement panel of the truck, while our model emphasizes on the wheels, which resulting in low relative risk scores. Meanwhile the Relative Risk scores are high in the image of a crane, as the two methods agree on where the important image patches are.
    }
    \label{fig:aim_3_rr}
\end{figure}

Figure \ref{fig:aim_3_rr} presents images on which we plotted contours reflecting TranSalNet's salience (orange) and alignment score heatmaps (blue) to visualize their overlap. For the two images on the left (bison and crane), the salience and alignment maps consistently show strong agreement across all three thresholding levels. For the two right images, there is no overlap. Specifically, the monkey's facial features are highly salient, but are not identified as important for alignment. In the case of the truck image, the banner area depicting colorful peppers is identified as salient, but the wheel area is identified as important for alignment. This is reasonable, as means of transportation in the set are effectively compared by observing the lower section of the vehicle, which differentiates trucks, cars, buses, motorcycles, trains and so on. Indeed we find these elements are often highly salient in the produced heatmaps. More results with appropriate level of detail are shown in the Appendix section below.

% \footnote{With approval of the organizing committee, this double anonymized link to figshare allows viewing more examples at appropriate level of detail: https://figshare.com/s/506d26037e071612cdf3 } 

% overlap figure

\subsection{Aim 4: Generalization to other architectures and training objectives}

\begin{figure}[tb]
    \centering
    \includegraphics[width=\textwidth, height=0.9\textheight]{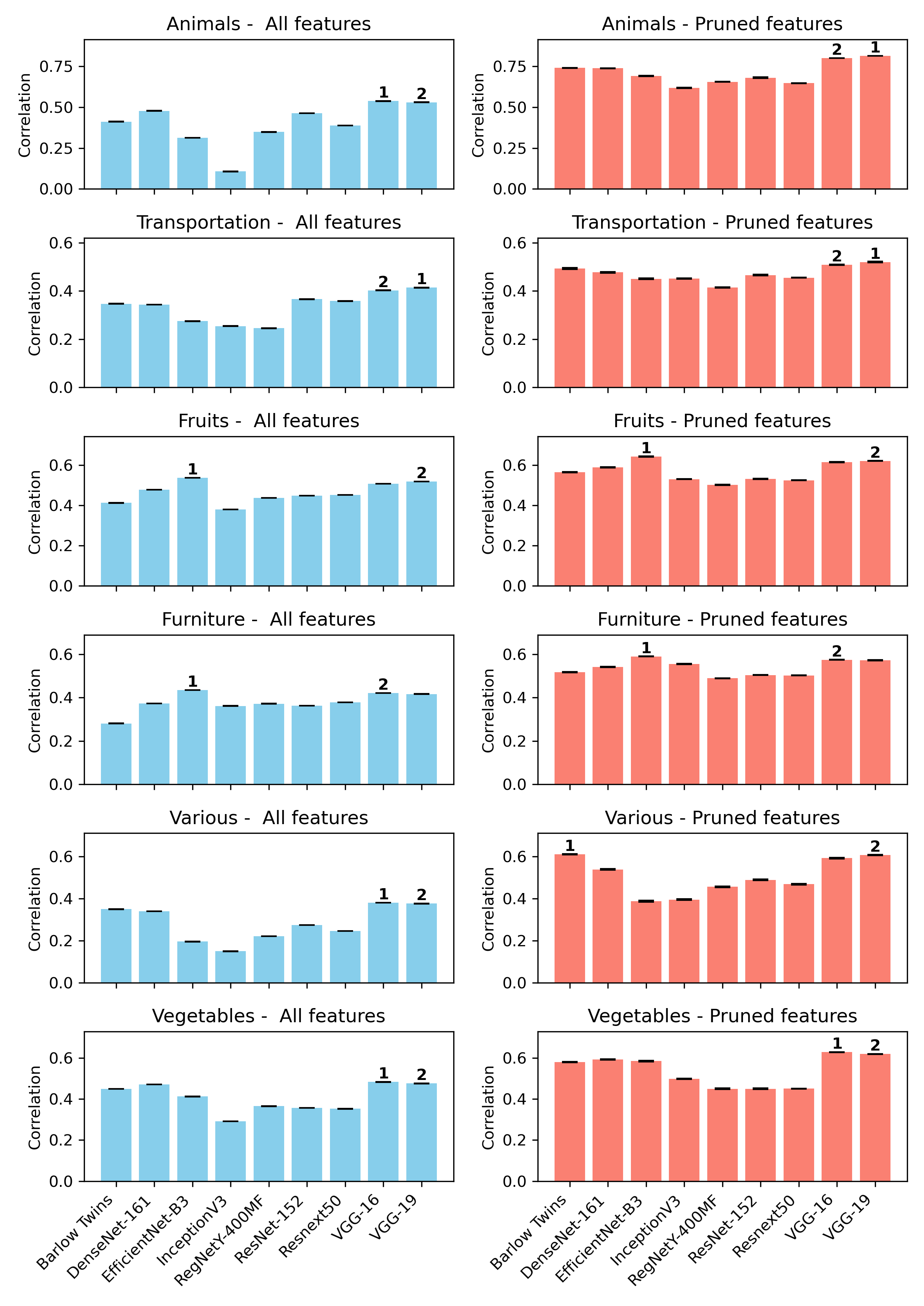}
    \caption{Cross-validation performance for models typically used as feature extractors. The numbers '1' and '2' refer to the two best performing models on the test set when using all features or only the features retained from the training set ('Pruned features').}
    \label{fig:aim_4_allcv}
\end{figure}

We find that quantifying alignment importance improved out-of-sample prediction of human similarity judgments across all architectures  and all six categories tested (see Figure \ref{fig:aim_4_allcv}).

Based on the results, we make the following observations. First, the two VGG-based architectures tended to perform the best overall, ranking first in three of the six image categories and second in all six categories. Second, baseline performance (test-set prediction using all features) tended to be diagnostic of which architecture would perform best using the learned pruned test set: for four of the six categories, the best performing model when using the full feature sets was also the best-performing when using the pruned sets. 

However, in three cases, none-VGG models predicted human judgments best. EfficientNet-B3 ranked highest for Fruits and Furniture. The compound scaling used in this architecture, which optimally balances width, depth and resolution has been argued to produce a better representation of relevant image details (see \cite{tan2019efficientnet} their Figure 7). Furthermore, as indicated in the Methods section, the fact that this model uses linear combinations of feature-map information for classification (after global pooling) makes it potentially more interpretable than VGG-16 and VGG-19, which use fully connected layers to learn complex combinations of feature-map information. Finally, the Barlow Twins architecture which is self-supervised and is not guided by a classification objective performed the best on the Various category. 

These findings suggest that the VGG architectures show considerable strength overall. However, the impact of removing single feature maps in these architectures is effectively evaluated via the changes in activations in the fully connected layers, which learn interactions between feature maps. Depending on the aims of the analysis, other architectures may be used if such interaction effects are of no interest. Practically, the findings of Aim 4 suggest that when using AIS-based heatmaps as explanations for human comparisons, it is is sensible to use an architecture that best predicts these judgments.

\section{Discussion}
Understanding what information is used in human comparisons is important not only for a better understanding of the comparison process itself, but also for comprehending how people form memories and make decisions \citep{roads2024modeling}. We introduced and validated a feature-map's Alignment Importance as a meaningful parameter relevant to such explanations. We first showed that AIS values generalize to improve prediction of human similarity judgments. This complements current approaches that achieve improvements by using reweighting or pruning of nodes in a DNN's penultimate layer \cite[e.g.,][]{peterson2018evaluating, attarian2020transforming,  kaniuth2022feature, jha2023extracting, tarigopula2023improved}.

We then used AIS to produce explanations for those judgments via heatmaps. These heatmaps offered some correspondence to state-of-the-art saliency maps, in that when saliency maps were thresholded at high percentiles, the resulting representation could sometimes predict (binarized) AIS heatmaps quite well, especially for Animals. However, instances where saliency and AIS-reduced maps diverged are of major theoretical importance as they show it is possible to dissociate visually salient image elements from those that are important for comparison.

Because the  method we present is based on mapping, or aligning a DNN's representational space to a human one via pruning, the feature space of the pretrained-DNN is of fundamental importance. For this reason, in Aim 1 we studied DNNs trained on both ImageNet and Ecoset datasets. We found that AIS scores improved out-of-sample prediction for models trained on either of the training datasets.  Thus, both models learn feature maps particularly relevant for accounting for the representational space of specific categories. For both Ecoset and ImageNet, category-specificity was shown in the fact that the relative ranking of AIS scores varied greatly across categories. Interestingly, Ecoset appears to distribute the AIS scores slightly more uniformly across feature-maps than ImageNet, which is a topic that requires further investigation. 

Further speaking to generalization across both training sets, the heatmaps were, for the most part, quite similar when created from Ecoset or ImageNet AIS scores, with average correlations between the heatmaps exceeding 0.75 for the Animals category.  However, some images showed low correlations, and these tended to be associated with more uniform post-softmax distributions in the DNN's categorization layer. This means that divergence in heatmaps produced by the two models were more prevalent for images that one of the models found difficult to classify. In practice, we recommend using both Ecoset and ImageNet trained models to create heatmaps and carefully evaluating images with inconsistent results.

The strongest demonstration of generalization of the AIS based approach was provided in Aim 4, where we showed that the method improves out-of-sample prediction of human similarity judgments across eight different architectures. From the perspective of construct validity, the choice of architecture is fundamental for the effective use of the proposed method. An architecture that provides poor out-of-sample predictions of human similarity judgments will  offer less meaningful explanations of human behavior compared to one that provides strong predictions.  Examining this issue we find that there was no architecture that provided the best prediction across all six image categories.  Thus, when explaining human comparisons for a stimulus set, it would be generally important to select an architecture with the best predictive capacity. 

However, we also note that predictive capacity should be considered conjointly with the complexity of the architecture. In the current study, we used the VGG architecture in Aims 1, 2 and 3, as it was the reference architecture in prior work on prediction of human similarity judgments from image embeddings \citep{attarian2020transforming, peterson2018evaluating, tarigopula2023improved}. As mentioned in the Methods and Results, the two VGG architectures, while providing good predictions, produce embeddings that naturally reflect interactions between feature map information, and so the removal of a feature map is assessed by the impact of its removal on these interaction values.  Other architectures that do not use fully connected layer after the deepest convolutions may produce simpler explanations. This is a topic that needs to be explored in future work.  

% Finally, we consider that the method can be further improved in the future to increase precision: since the current selection of feature maps is greedy, future works may incorporate heuristic information into the procedure to avoid this problem. Furthermore, quantifying the correlation among feature maps prior to computing AIS could potentially improve the assessment of their cognitive relevance compared to human data.



\clearpage

% \section*{Declarations}
 
% \subsection*{Ethical Approval} 
% Not applicable: No human data was collected in this study. All behavioral data used was previously collected by \citet{peterson2018evaluating} and made available to us upon our request. No new ethical approvals were required.

% \subsection*{funding}
% Not Applicable.

% \subsection*{Availability of data and materials} The materials should be requested from the above-mentioned authors. The source code used for generating heatmaps based on image embeddings and corresponding human similarity judgments is publicly accessible at  https://github.com/tlmnhut/ais\_heatmap

\bibliography{iclr2024_conference}
\bibliographystyle{iclr2024_conference}

\clearpage
\appendix
\counterwithin{figure}{section}
\section{Appendix}
\label{sec:appendix}

\subsection{Histogram of image-level correlations between Ecoset and ImageNet produced AIS maps}
\begin{figure}[h]
    \centering
    \includegraphics[width=0.7\textwidth]{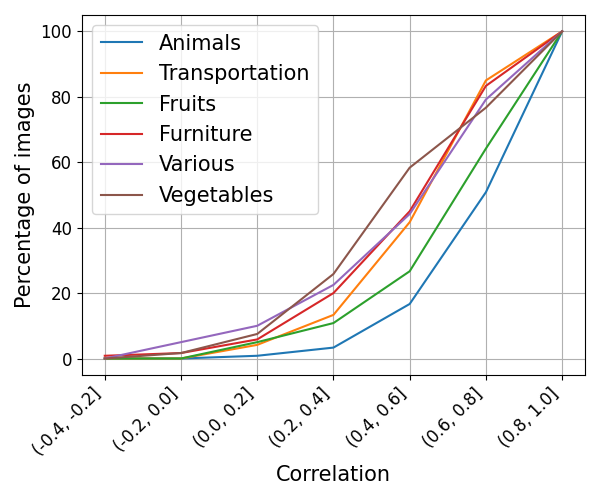}
    \caption{Cumulative histogram of correlations between heatmap' values created by ImageNet-trained and Ecoset-trained models.}
    \label{fig:aim_2_hist_heatmap}
\end{figure}

\newpage
\clearpage

\subsection{TranSalnet and AIS maps: Additional Images}
Additional images showing the overlap between the heatmaps created by Alignment Importance Scores (blue contours) and the saliency maps from TranSalNet (orange contours). Contours indicate the 5\%, 10\%, and 15\% most important pixels, with increasing color intensity respectively. Relative Risk values computed from  top 5\%, 10\% and 15\% pixels in each map are printed on the top of each images.

\begin{figure}[h]
    \centering
    \includegraphics[width=\linewidth]{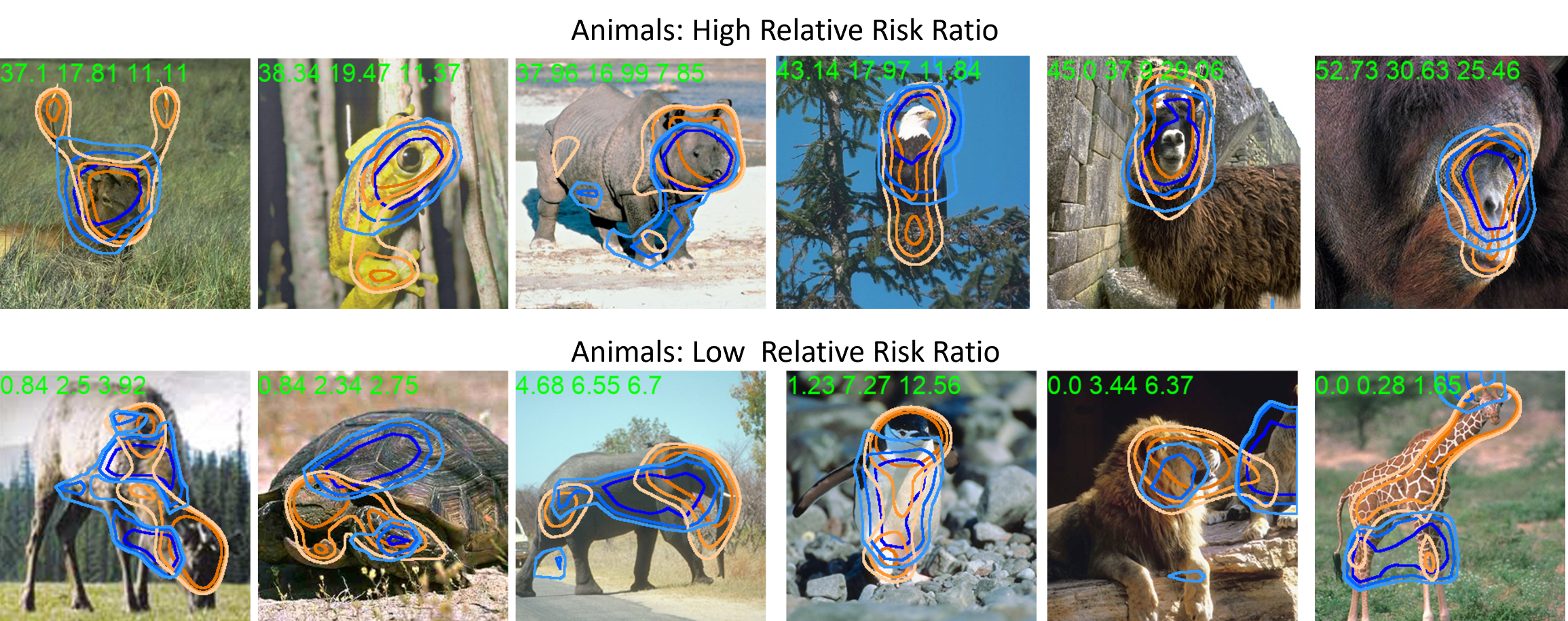}
    % \caption{..
    % }
    \label{fig:appendix_1}
\end{figure}

\begin{figure}[h]
    \centering
    \includegraphics[width=\linewidth]{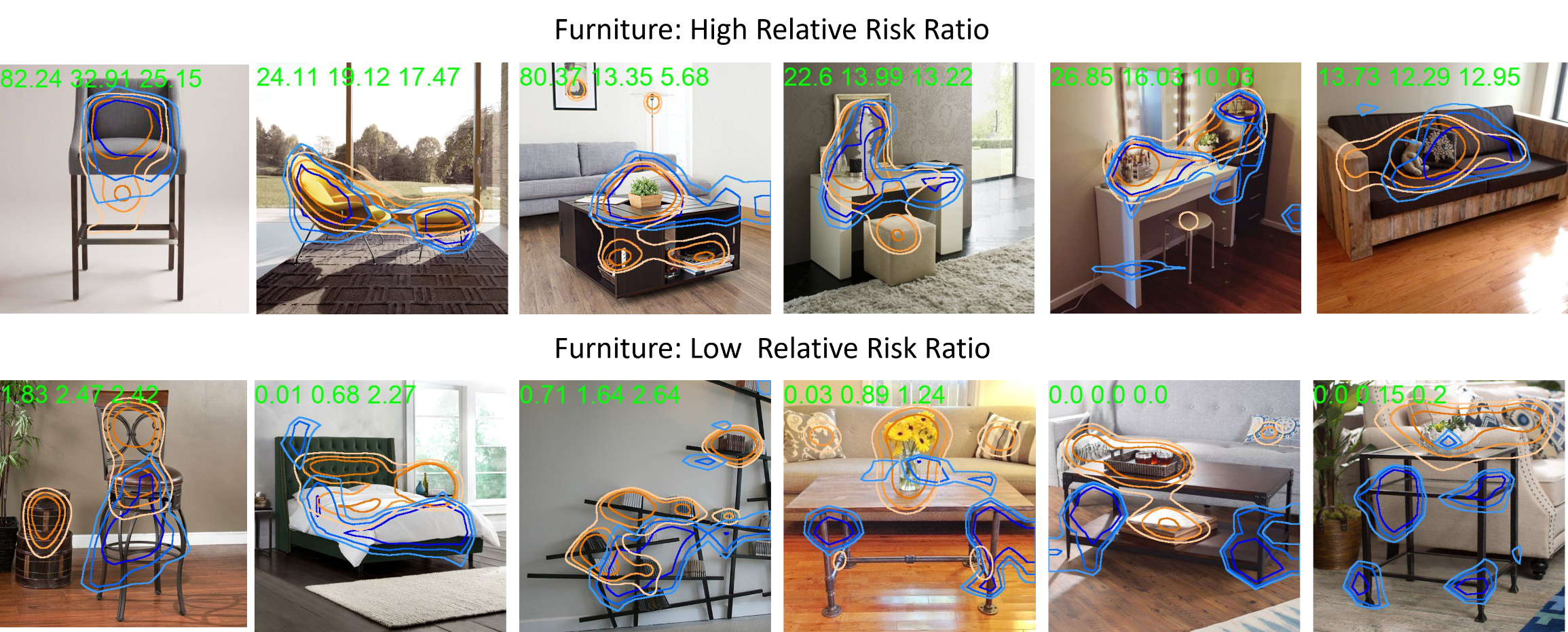}
    % \caption{..
    % }
    \label{fig:appendix_2}
\end{figure}

\begin{figure}[h]
    \centering
    \includegraphics[width=\linewidth]{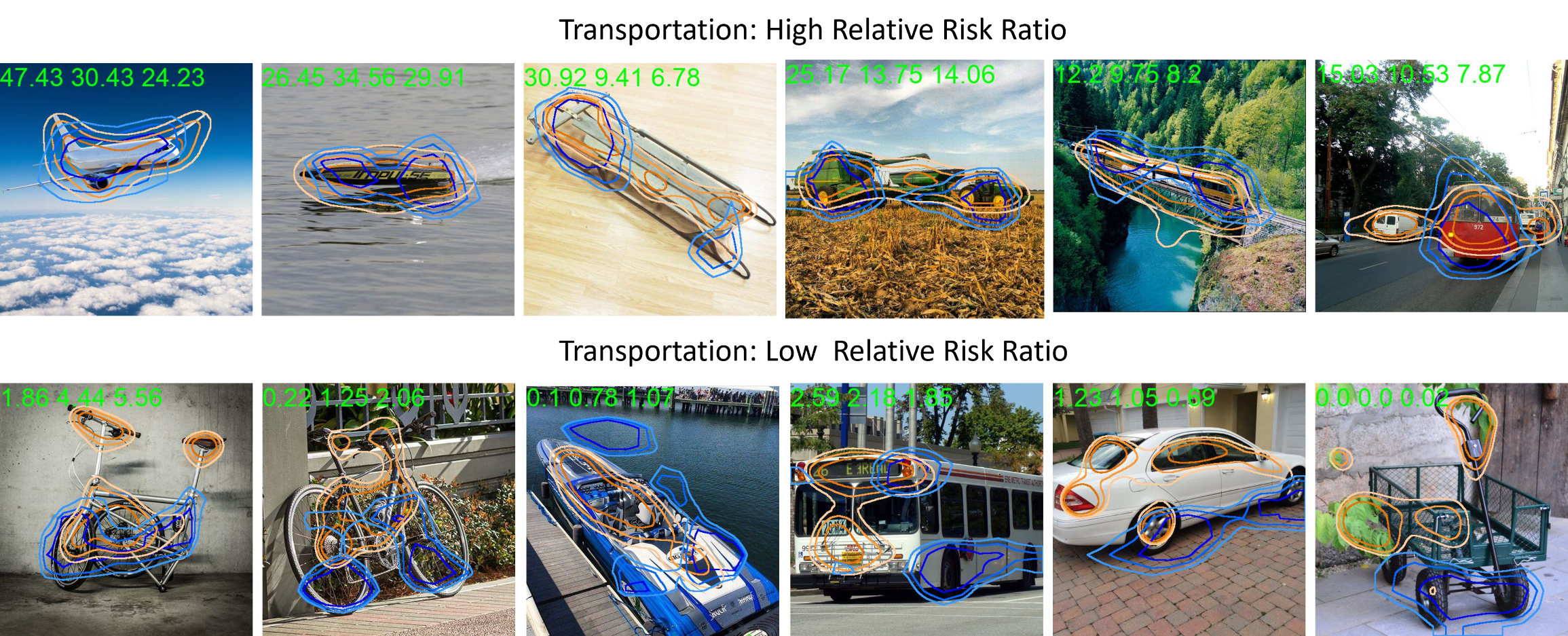}
    % \caption{..
    % }
    \label{fig:appendix_2}
\end{figure}

\newpage
\clearpage

\subsection{TranSalNet performance}
The image, below, adapted from Lou et al. (2022) shows performance of Translanet in prediction of human gaze. The figure presents the original image, the human gaze location (Ground Truth), and the gaze predictions made by Translanet, when trained on two different vision models.

\begin{figure}[h]
    \centering
    \includegraphics[width=0.9\linewidth]{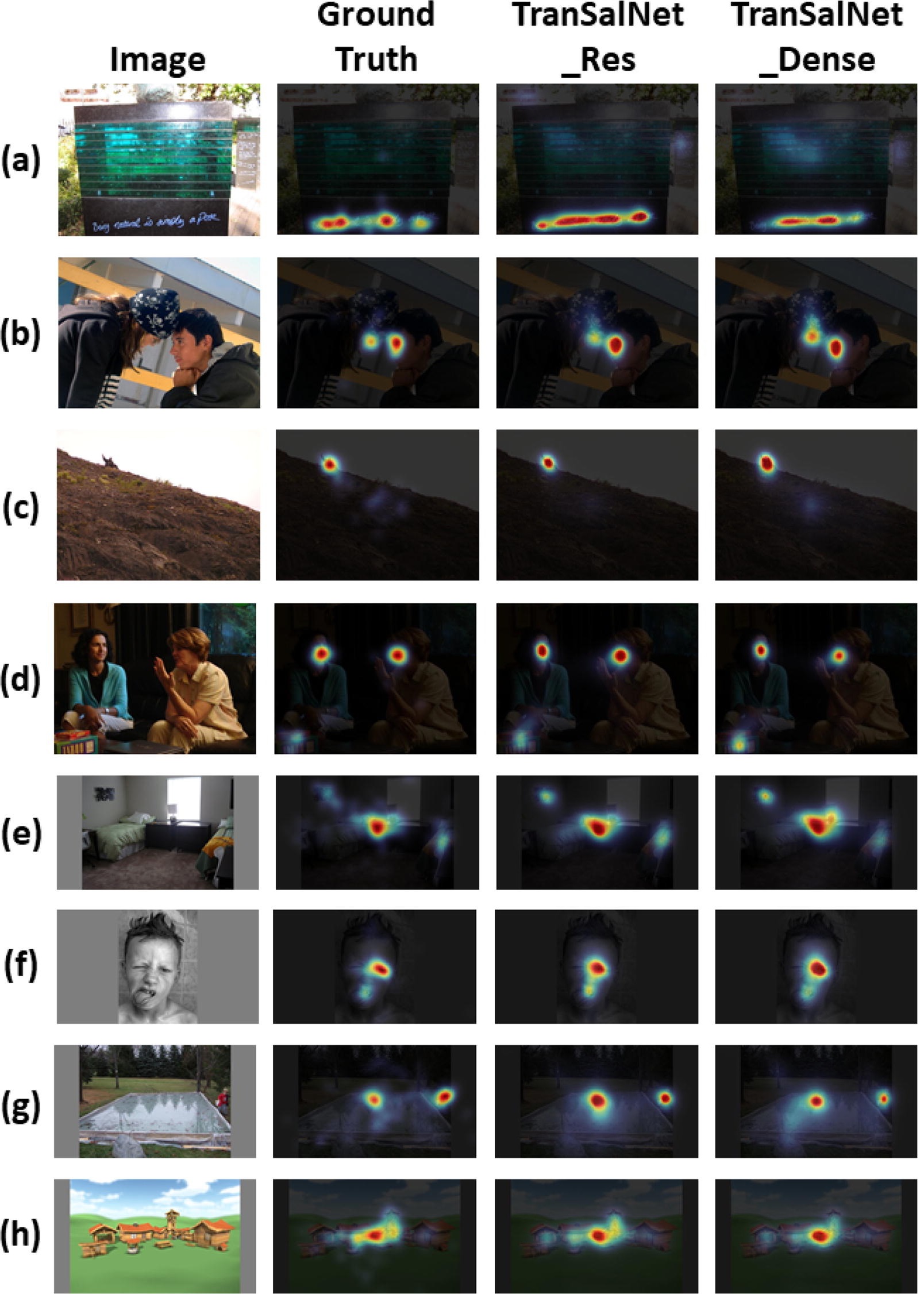}
    \caption{Figure adapted from Lou et al. (2022). https://doi.org/10.1016/j.neucom.2022.04.080. Original figure licensed CC-BY.}
    \label{fig:appendix_trans}
\end{figure}

\newpage
\clearpage

\subsection{Precision-Recall curves for Ecoset-produced images}
\begin{figure}[ht]
    \centering
    \begin{minipage}{0.49\textwidth}
        \centering
        \includegraphics[width=\textwidth]{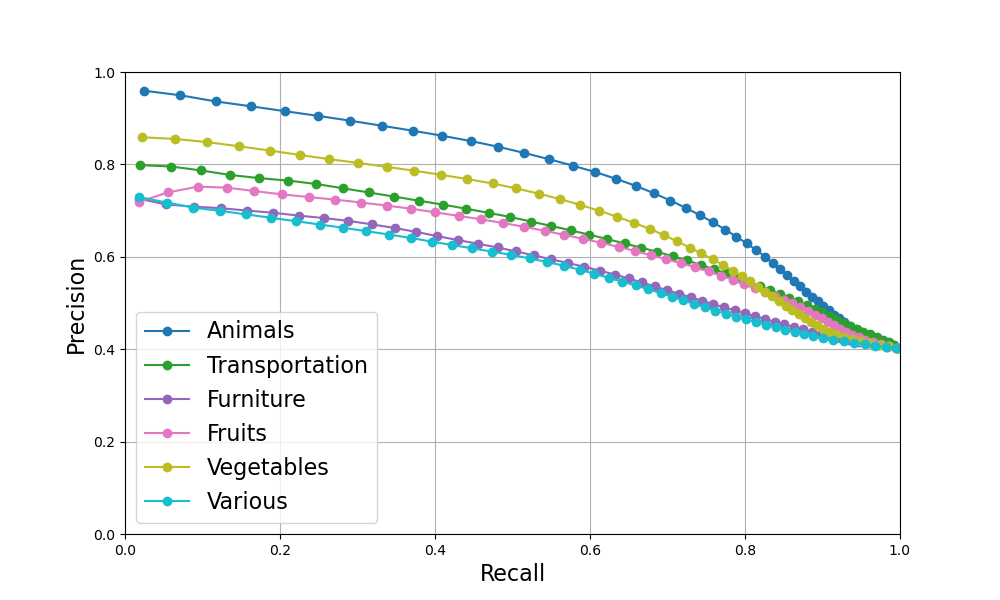}
        \subcaption{Target thresholded at 60th percentile}
        \label{fig:pr_curve_60_ecoset}
    \end{minipage}\hfill
    \begin{minipage}{0.49\textwidth}
        \centering
        \includegraphics[width=\linewidth]{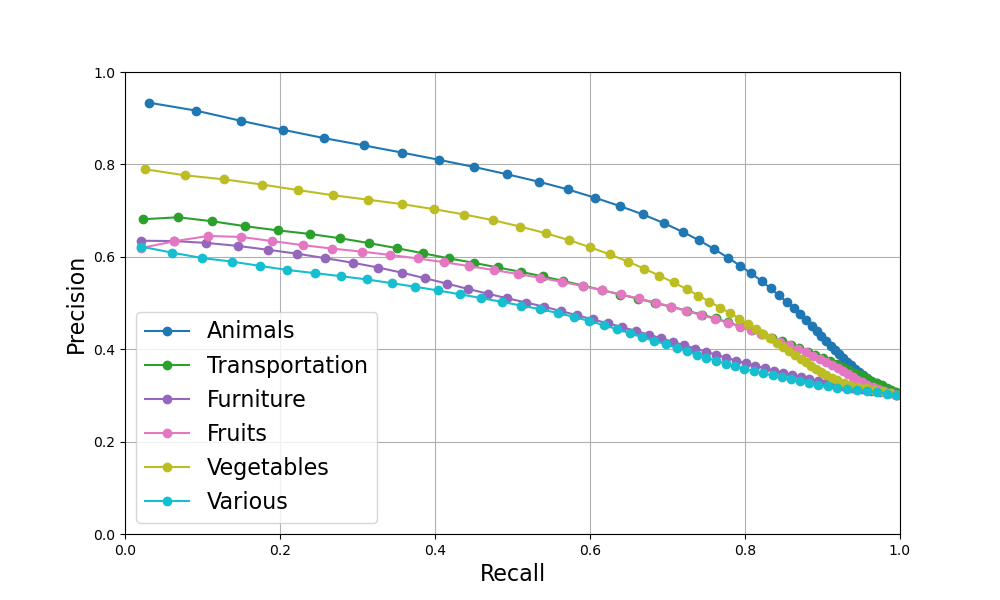}
        \subcaption{Target thresholded at 70th percentile}
        \label{fig:pr_curve_70_ecoset}
    \end{minipage}
    
    \vspace{0.5cm}
    
    \begin{minipage}{0.49\textwidth}
        \centering
        \includegraphics[width=\textwidth]{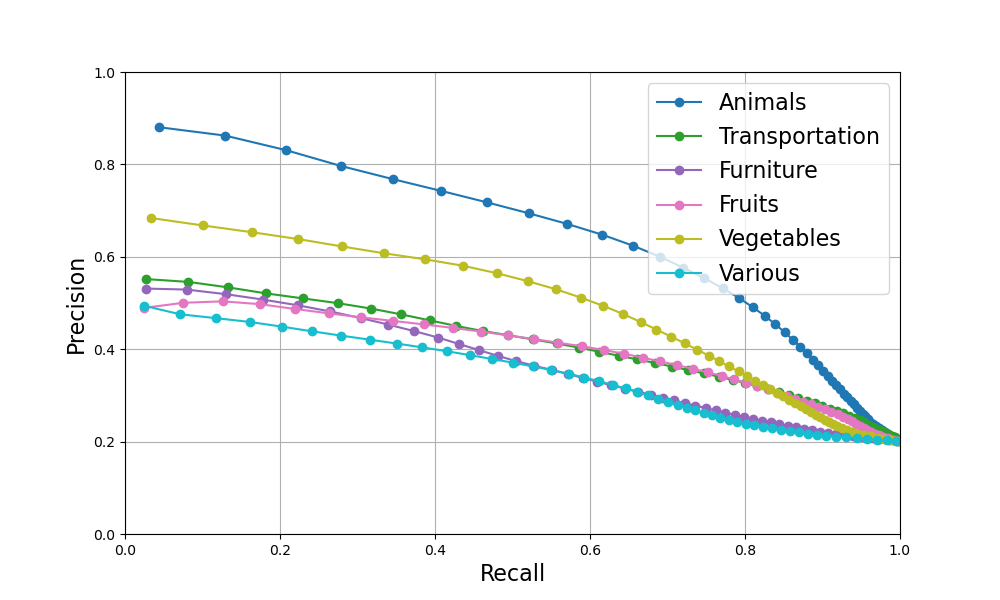}
        \subcaption{Target thresholded at 80th percentile}
        \label{fig:pr_curve_80_ecoset}
    \end{minipage}\hfill
    \begin{minipage}{0.49\textwidth}
        \centering
        \includegraphics[width=\textwidth]{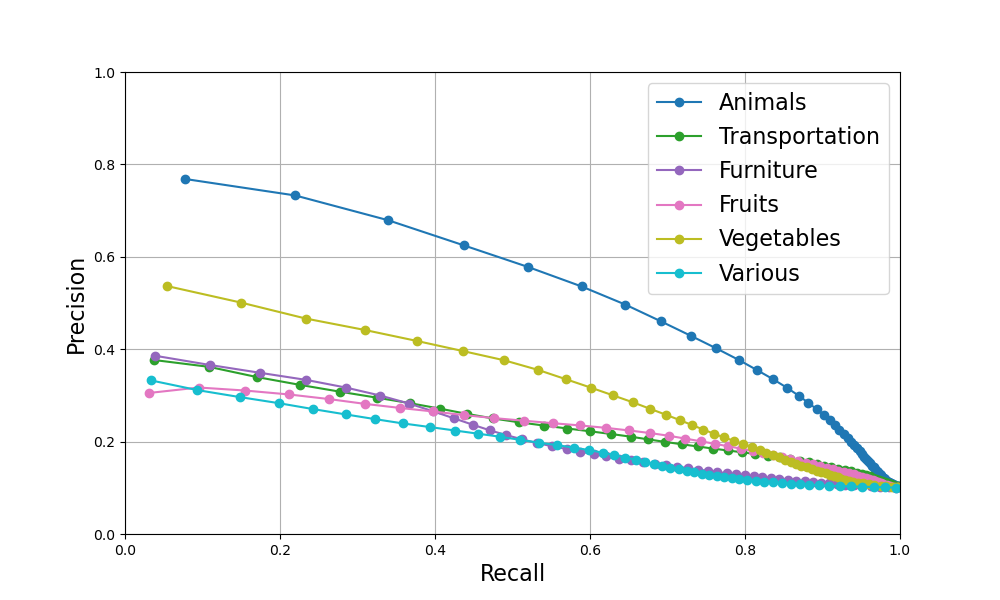}
        \subcaption{Target thresholded at 90th percentile}
        \label{fig:pr_curve_90_ecoset}
    \end{minipage}

    \caption{Precision-Recall Curves for different thresholds of the target variable. The target variable was heatmap values produced from AIS scores computed from Ecoset training. The predicting variable were saliency map values from obtained from TranSalNet}
    \label{fig:ecoset_pr_curves}
\end{figure}

\newpage
\clearpage

\end{document}